\documentclass[runningheads]{llncs}
\usepackage{amssymb,amsfonts}
\usepackage{algorithmic}
\usepackage{textcomp}
\usepackage{graphicx}
\usepackage{textcomp}
\usepackage{xcolor}
\usepackage{xcolor}
\usepackage{amsmath}
\usepackage{algorithm}
\usepackage{float}
\usepackage[T1]{fontenc}

\usepackage{graphicx}
\begin{document}
\title{Towards Robust Artificial Intelligence: Self-Supervised Learning Approach for Out-of-Distribution Detection}
\titlerunning{Towards Robust Artificial Intelligence}
%
\author{Wissam Salhab\inst{1}
\and
Darine Ameyed\inst{2} \and
Hamid Mcheick\inst{3} \and
Fehmi Jaafar\inst{4}}
\authorrunning{Salhab et al.}
\institute{555 University of Quebec at Chicoutimi, Quebec, G7H 2B1, Canada\\
\inst{1}\email{wissam.salhab1@uqac.ca}\\
\inst{2}\email{dameyed@uqac.ca}\\
\inst{3}\email{hamid\_mcheick@uqac.ca}\\
\inst{4}\email{fehmi.jaafar@uqac.ca}
}



\maketitle              
\begin{abstract}
Robustness in AI systems refers to their ability to maintain reliable and accurate performance under various conditions, including out-of-distribution (OOD) samples, adversarial attacks, 
and environmental changes. This is crucial in safety-critical systems, such as autonomous vehicles, transportation, or healthcare, where malfunctions could have severe consequences. This paper proposes an approach to improve OOD detection without the need of labeled data, thereby increasing the AI systems' robustness.
The proposed approach leverages the principles of self-supervised learning, allowing the model to learn useful representations from unlabeled data. Combined with graph-theoretical techniques, this enables the more efficient identification and categorization of OOD samples. Compared to existing state-of-the-art methods, this approach achieved an Area Under the Receiver Operating Characteristic Curve (AUROC) = 0.99.

\end{abstract}
\section{Introduction}
\label{ch:introduction} 
The rapid advancement of AI-driven systems has significantly impacted decision-making in various domains, including healthcare, finance, transportation, and security. However, AI systems often struggle with handling data from diverse sources, structures, and distributions, raising concerns about their robustness in real-world applications. Ensuring AI robustness is crucial, as these systems must consistently deliver reliable and accurate outputs under diverse conditions to maintain user trust and facilitate safe deployment and secure integration across multiple domains. This need is particularly pressing in high-stakes systems, where errors or incorrect decisions can have severe consequences and pose significant risk to human safety \cite{li2023trustworthy}.\\

A key aspect of AI robustness is the ability to detect OOD samples, thereby preventing unreliable predictions. Simply put, OOD occurs when input data differs from the training distribution. OOD samples can originate from various sources, often arising from unseen situations the model has not encountered during training. Moreover, they can also result from adversarial attack inputs deliberately designed to mislead the model. Consequently, understanding how OOD samples impact AI systems is essential to making them adaptable to dynamic input data.\\
Thus, ODD detection involves identifying inputs that differ significantly from the training distribution, including previously unseen classes, random noise, or adversarial attacks.
However, OOD detection remains a significant challenge. Many existing methods often struggle to accurately detect these outliers, especially when deep neural networks exhibit overconfidence, resulting in the misclassification of OOD data.\\

We propose a self-supervised method for extracting embeddings from unlabeled images using graph theory, which enables nuanced representations of both in-distribution and OOD data.\\
This study suggests a new OOD detection approach connecting self-supervised contrastive
learning, graph-based clustering, and Mahalanobis distance. Unlike previous methods, our approach is completely unsupervised, requires only in-distribution unlabeled data, which improves scalability and is targeted in real-world scenarios.\\
The integration of Graph clustering addresses the limitations of distance by modeling the distribution of data as a mixture of Gaussians, thereby improving the detection of near OOD samples. Compared to Maple \cite{venkataramanan2023gaussian}, which depends on label data and Gaussian assumption, our approach leverages the graph theory of capturing local data structures, and relative to to SSD \cite{sehwag2021ssd}, in contrastive learning but lacks clustering, our approach enhances robustness through graph representations, achieving superior AUROC score of 0.99 on benchmark datasets (CIFAR-10, CIFAR-100, SVHN).\\ 

The rest of this paper is structured as follows. Section \ref{ch:related_works} reviews related work in this field. Section \ref{ch:method} provides a detailed explanation of the proposed approach, followed by an experiment presented in Section \ref{ch:results}. The challenges and limitations are addressed in Section \ref{ch:threats}. Finally, we conclude the paper in Section \ref{ch:conclusion}, highlighting future research directions.

\section{Related Work}
\label{ch:related_works} 
OOD detection has been an active area of research. As we propose a self-supervised approach for ODD detection, this section reviews relevant related works in the field.
\subsection{Mahalanobis-Based OOD Detection}
\label{subsection:mahalanobis}
Mahalanobis distance (MD) is the distance between a data point and a distribution, and has been widely used in the literature for anomaly detection applications \cite{zhang2015low,ren2021simple}. In OOD detection, MD is frequently employed to distinguish far OOD samples from in-distribution samples, which differ greatly in meaning and style.\\
However, MD frequently fails to classify OOD samples close to the decision boundary and semantically similar to in-distribution samples \cite{NEURIPS2021_3941c435}. To address this limit, a straightforward solution known as the relative Mahalanobis distance (RMD) has been implemented as a more robust alternative. RMD enhances performance and increases resilience to hyperparameter fluctuations \cite{ren2021simple}. RMD relies on predefined statistical assumptions that often do not generalize well across diverse datasets or complex OOD situations. Its effectiveness also decreases in high-dimensional settings where estimating function correlations is difficult.
\\
\subsection{OOD Detection Using Supervised Classifiers} OOD classifiers help discover and classify data points that significantly deviate from a given data distribution \cite{mohseni2020self}. This is typically achieved by training a model using binary or multi-class classification approaches \cite{sun2022gradient}. \cite{venkataramanan2023gaussian} introduce MAPLE, which enhances OOD detection through predictive probability calibration. MAPLE dynamically clusters non-Gaussian class representations into multiple Gaussian distributions. The approach involves training a CNN using cross-entropy and triplet loss functions, combined with MD-based classification.\\
Their reliance on labeled data limits the effectiveness of many OOD detection methods, as they struggle to identify OOD instances that do not align with learned distributions. Additionally, the flexibility of these methods often relies on Gaussian assumptions, which may not always apply. This underscores the need for more adaptable OOD detection techniques that minimize the dependence on labeled data or rigid distribution assumptions. Contrastive learning models, for example, require large and diverse datasets to perform well and are sensitive to the selection of negative samples. Similarly, methods such as Maximum Mean Discrepancy (MMD) can be less effective in high-dimensional spaces, where distance metrics become unreliable. These issues underscore the necessity of developing robust OOD detection techniques that can effectively handle unlabeled datasets.
\subsection{OOD Detection Using Contrastive Learning} Self-supervised contrastive learning has gained significant attention, especially in computer vision and natural language processing tasks \cite{jaiswal2020survey}. The main objective of contrastive learning is to learn a representation of similar embedding in the feature space by grouping similar samples while pushing dissimilar ones apart \cite{tsai2020self}.\\
Sehwag et al. \cite{sehwag2021ssd} present SSD, a self-supervised system that utilizes unlabeled in-distribution data for ODD detection. SSD also incorporates training data labels and introduces enhancements for few-shot OOD detection.\\
Moreover, the authors of \cite{guille2022cadet} proposed CADet, a fully self-supervised anomaly detection method that leverages contrastive learning and the Maximum Mean Discrepancy (MMD) metric. CADet
examines whether two independent sample sets originate from the same distribution by combining contrastive learning with MMD, utilizing mean-based analysis. 

\subsection{OOD Detection Using Graph}
Liu et al. \cite{liu2023good} proposed a new approach for detecting OOD samples in graph-structured data. Their primary objective is to identify OOD graphs using unlabeled in-distribution data. They utilized the concept of contrastive learning to identify OOD graphs without the need for ground-truth labels \cite{liu2023good}. Whereas, other methods make use of Graph neural networks (GNN) to detect OOD samples within graph data \cite{gnn}.\\
Although these methods adapt OOD detection to graph structures, they face several challenges. GNN-based methods, in particular, require extensive tuning and often struggle with the complexity of high-dimensional graphs. Additionally, they often encounter scalability issues when handling large, densely connected graphs.\\
\subsection{Limitations of Previous Methods}
In this paper, we employ contrastive learning to extract embeddings from images in a self-supervised manner, enabling OOD detection without relying on labeled OOD samples. The proposed approach offers numerous advantages, including its ability to effectively capture feature correlations and maintain robustness in high-dimensional spaces without requiring labeled data. We summarize the limitations of existing methods and highlight our contributions in Table \ref{tab:comparison}. By eliminating the requirement for labeled data, reducing dependence on the Gaussian assumption, and optimizing computational efficiency through KNN-based graph constructions, our method provides a scalable and robust solution to detect OOD.
\begin{table}[]
    \centering
    \begin{tabular}{c c c c}
    \hline
         Method & Labeled Data & Gaussian Assupmtion & AUROC (CIFAR10 vs SVHN)  \\ \hline
         MAPLE \cite{venkataramanan2023gaussian} & Yes& Yes & 0.996\\
         SSD \cite{sehwag2021ssd} & No & No & 0.996\\
         Deep Ensemble \cite{larsson2019fine} & Yes & Yes & -\\
         VIT\_B\_16 \cite{ren2021simple} & Yes & No & - \\
         Ours & No & No & 0.999\\\hline
    \end{tabular}
    \caption{Comparison of OOD Detection Methods}
    \label{tab:comparison}
\end{table}

\section{Proposed Approach}
\label{ch:method} 
\begin{figure*}[ht!]
    \centering
    \includegraphics[width= \textwidth,height = 0.4\textheight]{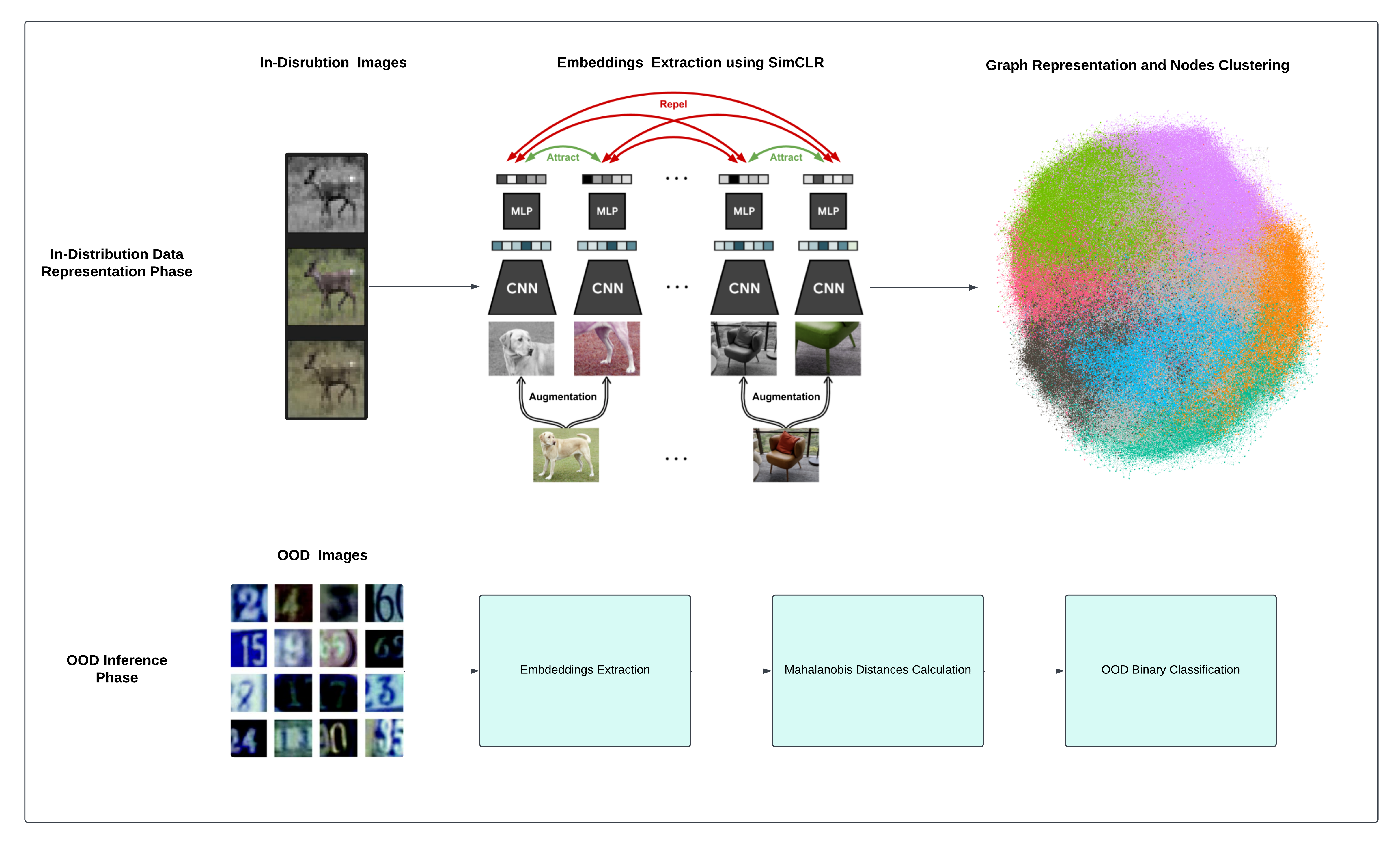} 
    \caption{The proposed approach includes two phases: \textbf{In-Distribution Data Representation Phase}: This phase involves extracting embeddings and creating a clustered graph based on these embeddings. \textbf{OOD Inference Phase}: This phase consists of extracting OOD embeddings, calculating the Mahalanobis distances to the in-distribution clusters, and performing binary classification for OOD detection.}
    \label{fig:GOOD}
\end{figure*}

In this section, we provide a detailed explanation of our proposed approach. We present the proposed approach in Figure \ref{fig:GOOD}. It consists of two phases: The In-Distribution data representation phase and the OOD inference phase. It provides a schematic overview of the proposed approach, clarifying the workflow from embedding extraction to OOD classification. These visualizations highlight the method’s ability to capture meaningful semantic structures, which supports its superior AUROC.


\subsection{Phase 1: In-Distribution Data Representation Phase}
This phase aims to group the in-distribution unlabeled samples into distinct clusters. It achieves this by extracting embeddings from input samples, constructing a representative clustered graph.

\subsubsection{Embeddings Extraction:}
We employ self-supervised contrastive learning, a practical approach that enables models to differentiate between augmented versions of the same image and other distinct images, thereby learning strong representations from unlabeled data \cite{sohn2016improved}.\\

Furthermore, SimCLR is a well-known framework that uses a contrastive learning approach  \cite{oord2018representation}. It simplifies contrastive self-supervised learning algorithms without the need for memory banks or specific architectures. SimCLR uses the infoNCE loss presented in Equation (\ref{equation:infoNCE}):\\
\begin{equation}
\label{equation:infoNCE}
\ell_{i,j} = -\log \frac{{\exp({\operatorname{sim}}}(z_i, z_j) / \tau)}{\sum_{k=1}^{2N} 1_{[k \neq i]} \exp({\operatorname{sim}}(z_i, z_k) / \tau)}\end{equation}\\
\begin{equation}
\label{equation:sim2}
= -{{\operatorname{sim}}}(z_i, z_j) / \tau + \log \left[\sum_{k=1}^{2N} 1_{[k \neq i]} \exp({\operatorname{sim}}(z_i, z_k) / \tau)\right]
\end{equation}
\\
where the batch size is represented as ${N}$, and ${\tau}$ represents the temperature parameter, ${z_i}$ and ${z_j}$ are two enhanced perspectives of the same case, and ${\operatorname{sim}(.,.)}$ represents the similarity function between two examples. The similarity between these embeddings is then calculated using cosine similarity, which forms the basis of the loss function.\\
In this study, we use SimCLR to extract image embeddings. SimCLR outperforms supervised methods when fine-tuned with a limited amount of labeled data \cite{sehwag2021ssd}.\\
However, embedding can be high-dimensional and still contain redundant information, leading to higher computation cost and model performance degradation \cite{liu2011latent}. After extracting the high-dimensional embedding, we employ Principal Component Analysis (PCA) to represent the latent space in lower dimensions, which reduces computational costs and mitigates performance degradation.



\subsubsection{Graph Representation and Nodes Clustering: }

Once the embeddings are extracted, we construct the representation graph, where each node represents an embedding and is connected to all other nodes. To simplify the graph, we remove nodes that have no edges. However, this method presents computational challenges, as a fully dense graph is inefficient and costly. To address this issue, we employ K-nearest neighbors (KNN) to reduce the number of edges in the graph, allowing each node to connect only to its nearest neighbors. The edges of the graph are weighted using cosine similarity as follows:\\
\begin{equation}
\label{equation:sim}
\text{{cosine similarity}} = \frac{{E_i \cdot E_j}}{{\|E_i\| \|E_j\|}}
\end{equation}
${E_i \cdot E_j}$ represents the dot product of the two embedding vectors ${E_i}$ and ${E_j}$, $\|E_i\|$ and $\|E_j\|$ stand for magnitude of the two vectors.\\
After constructing the graph, the Louvain method was used to group nodes within a graph into distinct clusters. This approach leverages modularity optimization to uncover the underlying structure of the graph, making it especially valuable for analyzing complex relationships and clusters \cite{combe2015louvain}.

\subsection{Phase 2: OOD Inference:}
After identifying the clusters in phase 1, phase 2 focuses on classifying new samples as in-distribution or OOD samples. In this phase, we extract embeddings from new samples and calculate Mahalanobis distances from each embedding point to the centroids of the clusters identified in phase 1. This involves binary classification of data points that exhibit significant deviations from the established in-distribution data.

\subsubsection{Embeddings Extraction:}
After we train the SimCLR model on the dataset, we extract embeddings from the OOD data during the inference phase. This method enables us to capture patterns and features without requiring labeled data, and it facilitates the calculation of Mahalanobis distances to the centroids of in-distribution clusters.
\subsubsection{Mahalanobis Distances Calculation:}
Unlike Euclidean distance, MD considers the covariance between the variables as shown in Equation (\ref{equation:mahalanobis}). In this study, $md$ is the distance between an input and the nearest centroid of the in-distributoin clusters.
\begin{equation}
\label{equation:mahalanobis}
    md = \sqrt{(\mathbf{x} - \mathbf{\mu})^T \mathbf{S}^{-1} (\mathbf{x} - \mathbf{\mu})}
\end{equation}
$\mathbf{x}$ is the input vector, $\mathbf{\mu}$ is the centroid of the cluster, and $\mathbf{S}$ is the covariance matrix of the cluster. We use MD to calculate the distance of the samples from the centroid of the in-distribution clusters.
\subsubsection{OOD Binary classification:}
While a threshold is commonly used for practical classification, it is essential to differentiate between its role in decision-making and overall performance evaluation. Following the approach outlined in \cite{sastry2020detecting}, the authors demonstrated the effectiveness of setting the threshold at the 95th percentile of distances within the in-distribution dataset. In our study, we also set the threshold at the 95th percentile of the in-distribution data, solely to evaluate classification accuracy within this dataset. Although this specific threshold helps assess in-distribution classification accuracy, we use the AUROC as the primary evaluation metric for OOD detection. Unlike threshold-based methods, the AUROC is threshold-independent and measures the classifier's ability to distinguish between in-distribution and OOD data across various thresholds. This provides a more comprehensive assessment of the Mahalanobis distance-based OOD classifier.\\
\subsection{Theoretical Justification for Graph Clustering and Mahalanobis Distance}
The proposed approach is integrated to enhance contrastive learning and graph-based clustering for OOD detection, utilizing the Mahalanobis distance. This section provides a theoretical foundation for why graph clustering enhances Mahalanobis-based OOD detection. Graph clustering, achieved through the Louvain method, involves grouping different in-distribution embeddings into distinct groups based on their latent space similarity. This process captures the underlying structure of the data and addresses the limits of the Mahalanobis distance in high-dimensional settings. The Mahalanobis distance is considered for a single Gaussian distribution of data, which may fail when the data exhibits multi-model or non-Gaussian characteristics \cite{NEURIPS2021_3941c435}. Concerts we implement, effectively modeling assumptions as a mixture of Gaussians, where each cluster represents a local Gaussian component. This means that the Mahalanobis distance is calculated with respect to the nearest cluster centroid, which improves the sensitivity of OOD tests near the decision limits. By reducing $md$ in Equation \ref{equation:mahalanobis} on all clusters, we choose the nearest cluster, which improves the strength of distributional shifts. The Louvain method optimizes modularity, ensuring that clusters are closely connected while being externally connected, and interact with the perception that they disassemble the data into units. This grouping reduces the effect of high-dimensional covariance estimation errors, as $S$ is calculated on small, more homogeneous groups. In addition, the use of KNN reduces the graph calculation complexity by limiting the connection to the nearest neighbors, which minimizes the graph from $O(n^2)$ to $O(n^K)$. These savings preserve the local neighborhood.

\section{Experimental Results}
\label{ch:results}
Our primary results confirm the effectiveness of our proposed approach. Across several datasets, we observe a significant increase in the AUROC, achieving a maximum value of 0.999.
\label{section:experiment}
\subsection{Datasets}
We validate our approach using well-used datasets in computer vision and machine learning, including CIFAR10, CIFAR100, and SVHN.\\
\textbf{CIFAR10}: consists of 60,000  images classified into 10 classes, each containing 6,000 photos. It includes 10,000 test images and 50,000 training images. This dataset is a commonly used resource for training computer vision and machine learning algorithms. In this study, no labeled training images were used.\\
\textbf{CIFAR100}: This dataset comprises of 60,000 photos, each measuring 32x32 pixels, divided into 100 classes with 600 photos in each class. It includes 50,000 training images and 10,000 test images. The training images without labels were used in this study.\\
\textbf{SVHN}: This dataset includes 600,000 images of house numbers from Google Street View, including 73,257 training images and 26,032 images as the testing dataset \cite{svhn}. The training images without labels were used in this study. The code and results of this study are available on GitHub \cite{studyURL}.

\subsection{Experiment Setup}
The SimCLR model is trained using a batch size 256 to extract image embeddings. The temperature parameter is set to ${\tau}$ = 0.1, and the learning rate to 0.0001. A weight decay of ${\mathbf{1e^{-4}}}$ is applied and the training lasts over 300 epochs. We utilize 4 GPUs, including an NVIDIA 8GB V100, which completes the training in 8.795 hours. The process involves transforming the input images using various techniques, including Random Horizontal Flip, Random Resized Crop, Color Jitter, and Random Grayscale. These transformations adjust the images' brightness, contrast, saturation, and hue. The ResNet50 architecture serves as the base encoder, with an output dimension of 2048. The final layer of this model consists of a multi-layer perceptron (MLP), which includes a linear layer followed by a rectified linear unit (ReLU) activation function. The last layer functions as a projection layer with a dimension of 128. Before extracting features for the downstream task, we remove the projection head from the model. KNN was used for graph construction; we used k = 7 to balance cluster accuracy and computational efficiency (see Tables 2 and 3).

\subsection{Results}
\subsubsection{Embeddings Extraction:} 
\begin{figure}[h!]
    \centering
    \includegraphics[scale=0.8]{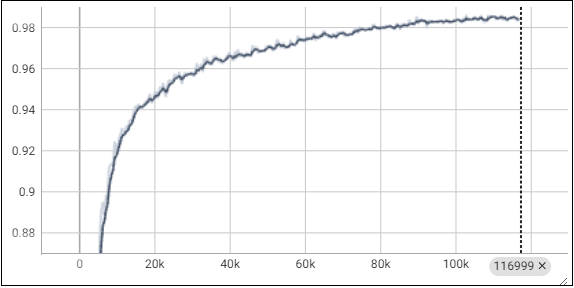}
    \caption{Top-5 validation accuracy chart. The model reaches a top-5 validation accuracy of 98.2\% after 117000 steps for the CIFAR10 dataset. A top-5 accuracy indicates that the true labels are among the five highest predicted probabilities.}
    \label{fig:valtop5}
\end{figure}

The result of SimCLR training with this setup is shown in Figure \ref{fig:valtop5}, which indicates an accuracy of 0.982 for top-5 validation on the CIFAR10 dataset. After extracting embeddings, we utilize PCA to reduce dimensionality. Following the SimCLR training on CIFAR-10, we leverage the learned representations to extract embeddings for all other datasets. 
These embeddings are then utilized to evaluate the model's performance on downstream tasks, specifically clustering and OOD detection. The results confirm that these learned representations effectively capture meaningful semantic information despite domain shifts between datasets.
\subsubsection{Graph building and clustering:}
\begin{figure}[]
    \centering
    \includegraphics[scale = 0.35]{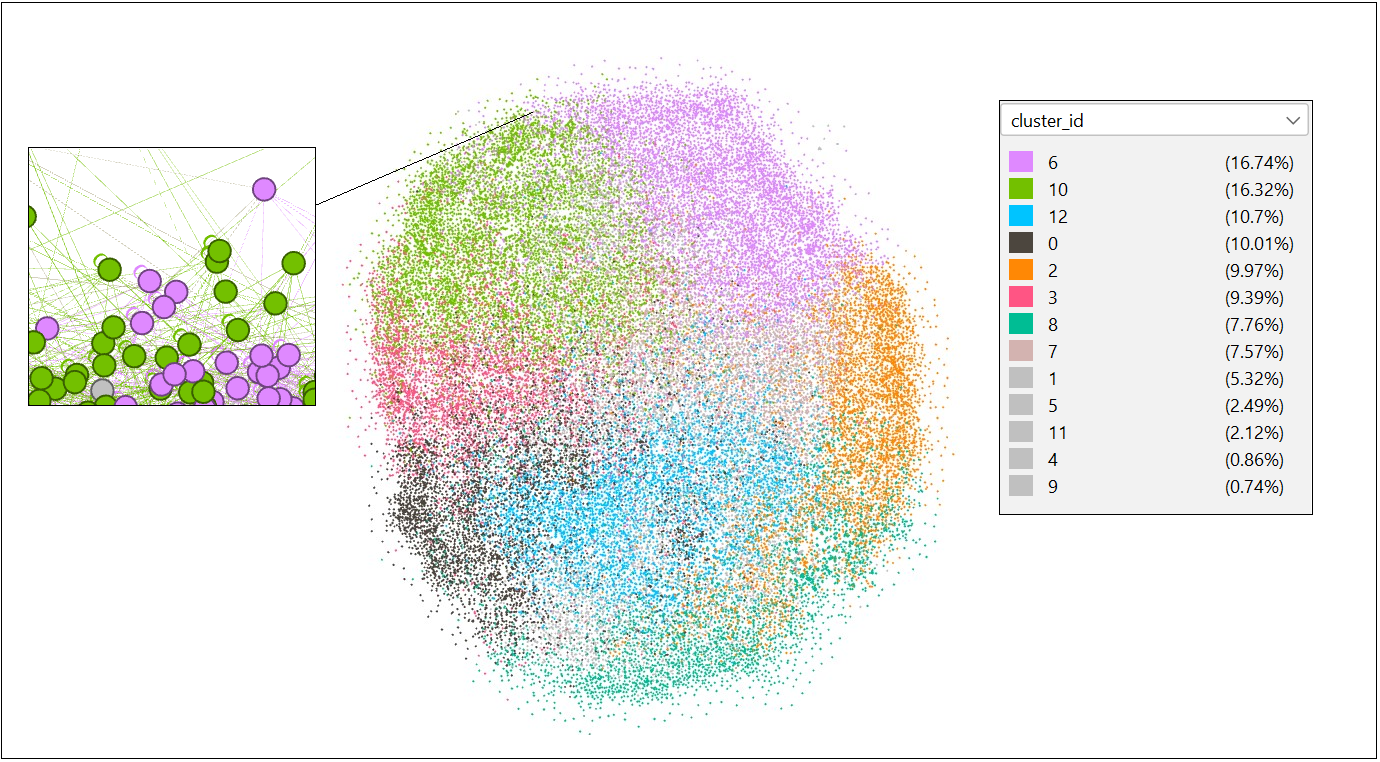}
    \caption{Clustered Graph: Each node in the graph represents an image embedding in latent space. We obtained 13 clusters (differentiable via colors in this image) out of 10 clusters.}
    \label{fig:gl_k7}
\end{figure}
Choosing 'k' neighbors in KNN is essential as it establishes the graph structure. Instead of calculating and storing the full adjacent matrix, we connect each data point to its closest neighbors. This graph reduces the calculation costs and the memory required to store the graph representation. This approach provides a sufficient efficiency advantage by reducing the density of the edges. \\
After creating the graph, we use the Louvain method to identify clusters within it. This method is designed to recover non-overlapping communities from large networks \cite{Loulou}. It involves calculating the degree to which the nodes in a cluster are more densely linked than they would be in a random network. Figure \ref{fig:gl_k7} shows us the graph after performing Louvain clustering.\\
Moreover, we test the number of neighbors to consider in the KNN with several values of 'k'. The findings for CIFAR100 are shown in Table \ref{table:k7v100}.
When the number of neighbors 'k' increases from 5 to 9, the number of edges rises from 180480 to 335785, while the number of clusters decreases from 167 to 90. Overall, the graph presents a total of 109 clusters at k = 7, which is close to the target of 100 clusters, the ground truth in the case of CIFAR100. At k = 7, the graph contains 258,385 edges and 50,000 nodes.
The results for the CIFAR10 graph are presented in Table \ref{table:k7v10}. As the number of neighbors, 'k', increases from 5 to 11, the number of clusters first decreases from 16 to 13. However, when 'k' reaches 11, the number of clusters increases to 14.
\begin{table}[ht]
\caption{Using different k-values for KNN \textbf{with cifar100} dataset, The optimal value of k is 7, resulting in 109 clusters, the closest to the ground truth in CIFAR100 clusters.}
\label{table:k7v100}
\centering
\begin{tabular}{|c c c c|}
\hline
K & Nodes & Edges & Clusters \\
\hline
k=5 & 50000 & 180480 & 167 \\
\hline
k=7 & 50000 & 258385 & 109 \\
\hline
k=9 & 50000 & 335785 & 90 \\
\hline
\end{tabular}
\end{table}
The selection of k = 7 resulted in a graph with 180,451 edges, 50,000 nodes, and 13 clusters, which is closest to 10, the ground truth in CIFAR-10. In contrast, k = 9 had 335,773 edges and the same number of clusters (13), but with more edges, resulting in a higher computational cost.\\
\begin{table}[h!]
\caption{Using different k-values for KNN with \textbf{cifar10} dataset, The optimal value of k=7 results in 13 clusters, closest to the ground truth of CIFAR10 clusters.}
\label{table:k7v10}
\centering
\begin{tabular}{|c c c c|}
\hline
K & Nodes & Edges & Clusters \\
\hline
k=5 & 50000 & 258280 & 16 \\
\hline
k=7 & 50000 & 180451 & 13 \\
\hline
k=9 & 50000 & 335773 & 13 \\
\hline
k=11 & 50000 & 412678 & 14 \\
\hline
\end{tabular}
\end{table}
\subsubsection{OOD Detection Results: }
We present the results of the OOD detection in Table \ref{table:cifar10vs100}. When we train the model on CIFAR10 (In-Distribution) and test it on CIFAR100 and SVHN (OOD datasets), our proposed approach achieves nearly the best AUROC and AUPR scores of 0.999 on both OOD datasets, significantly outperforming MAPLE \cite{venkataramanan2023gaussian}, Deep Ensemble \cite{larsson2019fine}, SSD \cite{sehwag2021ssd}, and even the competitive Vision Transformer (ViT-B16) \cite{ren2021simple}. 
\begin{table*}[hbt!]
\centering
\caption{The performance of our approach compared to existing methods on the same datasets: \textbf{CIFAR10 (In-Distribution), CIFAR100 and SVHN (ODD).}}
\label{table:cifar10vs100}
\begin{tabular}{|c|c|cc|cc|}
\hline
\textbf{Method} & \textbf{ID Accuracy} & \multicolumn{2}{c|}{\textbf{OOD (AUROC)}} & \multicolumn{2}{c|}{\textbf{OOD (AUPR)}} \\

 & & \textbf{CIFAR100} & \textbf{SVHN} & \textbf{CIFAR100} & \textbf{SVHN}  \\
\hline
MAPLE \cite{venkataramanan2023gaussian} & 0.956 \textpm 0.01 & 0.926 \textpm 0.01 & 0.996 \textpm 0.01 & 0.918 \textpm 0.01& 0.997 \textpm0.01\\
     Deep Ensemble \cite{larsson2019fine} & 0.964  \textpm  0.01 & 0.864  \textpm 0.01 &-&0.885 \textpm  0.01&-\\
    SSD  \cite{sehwag2021ssd} &          -                &   0.906                                 & 0.996&0.892&-\\
    $SSD_k$(k=5)  \cite{sehwag2021ssd} &    -                       &  0.931 &0.997&0.919&-\\
    ViT-B\_16 \cite{ren2021simple} &-  &0.998 & -&-&-\\ 

    Our & 0.95 & 0.999& 0.999&0.999&0.999\\
\hline
\end{tabular}
\end{table*}
Similarly, when trained on CIFAR100 (In-Distribution) and evaluated on CIFAR10 and SVHN (out-of-distribution datasets), the proposed approach maintains exceptional performance with AUROC scores of 0.999 and 0.997 for CIFAR10 and SVHN, respectively (Table \ref{table:cifar100vs10}), again surpassing all comparative methods. The proposed approach consistently exhibits substantial improvements across both evaluation scenarios, underscoring its robustness and effectiveness in distinguishing OOD samples.
\begin{table*}[h!]
\caption{The performance of our approach compared to existing methods on the same datasets: \textbf{CIFAR100 (In-Distribution), CIFAR10 and SVHN (ODD).}}
\label{table:cifar100vs10}
\centering
\begin{tabular}{|c|c|cc|cc|}
\hline
\textbf{Method} & \textbf{ID Accuracy} & \multicolumn{2}{c|}{\textbf{OOD (AUROC)}} & \multicolumn{2}{c|}{\textbf{OOD (AUPR)}} \\

 & & \textbf{CIFAR10} & \textbf{SVHN} & \textbf{CIFAR10} & \textbf{SVHN}  \\
\hline
     MAPLE \cite{venkataramanan2023gaussian} &  0.789   \textpm  0.01                      &   0.793 \textpm 0.01      & - & 0.799    \textpm  0.01   &-                 \\
     Deep Ensemble  \cite{larsson2019fine} & 0.796  \textpm  0.01 & 0.798 &-&0.792 \textpm  0.01 & - \\
      SSD \cite{sehwag2021ssd} & -                        &      0.696                  &0.949           & 0.645&- \\
    $SSD_k$(k=5) \cite{sehwag2021ssd}  &    -                       & 0.782  &0.991&0.772&- \\
ViT-B\_16 \cite{ren2021simple} & - & 0.9442 & -&-& -\\ 
   Our & 0.97 & 0.999&0.997&0.999& 0.997\\\hline
\end{tabular}
\end{table*}
\subsection{Discussion}
The results of our proposed approach demonstrate significant advancements in addressing OOD detection challenges. Our method begins by constructing a graph from embeddings through self-supervised learning, eliminating the need for labeled data. This graph-based structure effectively captures the local neighborhood relationships between data points, providing a more nuanced understanding of the underlying manifold and enhancing the discriminative power for tasks such as clustering and OOD detection.\\ 

However, while the self-supervised approach significantly improves the incremental robustness of AI systems in detecting OOD, it is not without drawbacks. Its effectiveness heavily depends on the quality and diversity of the raw data. The integration of graph-based clustering proves highly beneficial, enabling more effective grouping of data points. By uncovering underlying structures within the dataset that traditional methods may overlook, our approach reveals patterns that can inform better decision-making in OOD detection scenarios.\\ 
Furthermore, leveraging Mahalanobis distance for similarity measurement with a graph enhances the robustness of our classification strategy. This metric provides precise distance calculations between data points and their respective clusters, enabling the model to maintain high classification performance even when encountering distributional shifts.\\

Overall, our findings underscore the practical applicability of the proposed approach. It offers a scalable and efficient solution tailored for real-world OOD detection challenges, paving the way for more reliable and practical machine learning applications across various domains.

\section{Challenges and Limitations}
\label{ch:threats}
OOD detection poses several challenges and limitations that hinder its implementation and efficiency in different applications. Understanding these barriers is crucial for developing more effective OOD detection methods.\\
One major challenge lies in the process of generating OOD samples, which remains poorly understood. This ambiguity creates significant obstacles in building reliable systems.\\
Another critical issue is setting an appropriate threshold for the in-distribution that effectively flags OOD examples without increasing the risk of false positives or false negatives. This remains challenging when dealing with unlabeled datasets. Calibrating a threshold that is both sensitive enough to detect OOD samples and precise enough to prevent misclassification is a complex task, especially when dealing with an unlabeled dataset. \\
Moreover, failing to detect an OOD input can lead to poor decision-making or system errors, which is particularly critical in safety-sensitive applications.
Conversely, the computational requirements for the OOD detection method increase with model complexity and the size of the data. For instance, the complexity Louvain method alone used in this study ranges from \( O(N \log N) \) to \( O(N^2) \), with memory usage reaching \( O(N^2) \) due to the adjacency matrix, posing scalability challenges for extra large datasets that are resource-intensive.\\
Moreover, the effectiveness of an OOD detection model is closely tied to the quality and diversity of the training dataset. This limitation highlights the need for comprehensive and representative datasets to enhance the model's ability to identify a wide range of OOD scenarios.

\section{Conclusion and Future Work}
\label{ch:conclusion}
We highlight the significant progress in enhancing the trustworthiness of AI systems, especially through advances brought by self-supervised learning. We delve into the challenges faced by AI and propose a novel approach for out-of-distribution detection, based on Mahalanobis distance, graph theory, and contrastive learning. We emphasize the critical role of OOD detection in ensuring the robustness of AI systems without the need for labeled data.\\
For future work, we plan to integrate our proposed approach with uncertainty estimation methods and existing AI systems, particularly those deployed in safety-critical applications. This integration aims to further enhance the trustworthiness of AI systems in real-world scenarios. Ensuring the robustness and reliability of AI systems is essential for their widespread adoption and long-term success across various industries.



\bibliographystyle{splncs04} 
\bibliography{refs}
\section{Ablation Study}
In this Section, we tried to isolate the effect of the clustering algorithm and confirm the importance of graph-based embedding representations.
\subsection{Supervised OOD detection using KNN without graph representation}
In this section, we implemented a direct neighbor (KNN) algorithm on raw data without performing any embedding extraction or graph representation. The purpose was to evaluate the performance of KNN in differentiating between the in-distribution data (CIFAR-10) and the OOD (CIFAR-100) data and to emphasize the role of embedding extraction on OOD classification. We used values ranging from 1 to 14 with different K values and calculated the AUROC for each value. The consequences summarized in Table \ref{tab:auroc_scores} indicate that the AUROC scores improve but are still limited, reaching a maximum of 0.54 at K = 15 for CIFAR10 and CIFAR100. This indicates that many neighbors can provide more effective discrimination between the two classes in this context.
\begin{table}[h!] 
\centering 
\caption{AUROC Scores for Different Values of \( K \)} 



\begin{tabular}{|c|cc|cc|}
\hline
\textbf{k} & \multicolumn{2}{c|}{\textbf{CIFAR10(ID)}} & \multicolumn{2}{c|}{\textbf{CIFAR100(ID)}} \\

 neighbors& \textbf{} vs& \textbf{CIFAR100(OOD)} & \textbf{} & vs \textbf{CIFAR10(OOD)}  \\

\hline
5  & &0.5292 && 0.5292 \\
7  & &0.5319  && 0.5318\\
9  & &0.5360 && 0.5360\\
11 & &0.5381 && 0.5380 \\
13 & &0.5392 &&  0.5392\\
15 & &0.54219 && 0.5406\\ \hline 
\end{tabular} \label{tab:auroc_scores} \end{table}
\subsection{K-means for Node Clustering}
In this section, we attempted to implement K-means as the clustering algorithm instead of the Louvain method. However, the efficiency remains high, as demonstrated in Table \ref{table:kmeanscifar100vs10} while the in-distribution accuracy decreased from 0.97 to 0.95. The same holds for other datasets; the impact of the clustering algorithm was minimal. This confirms the robustness of our proposed approach, which relies on graph representation for data embedding to detect OOD instances.

    

\begin{table*}[h!]
\caption{The performance of our approach compared to existing methods on the same datasets: \textbf{CIFAR100 (In-Distribution), CIFAR10 and SVHN (ODD)}, using K-means where k=109 ( same k as the number of clusters obtained via the Louvain method)}
\label{table:kmeanscifar100vs10}
\centering
\begin{tabular}{|c|cc|cc|}
\hline
 \textbf{ID Accuracy} & \multicolumn{2}{c|}{\textbf{OOD (AUROC)}} & \multicolumn{2}{c|}{\textbf{OOD (AUPR)}} \\

  & \textbf{CIFAR10} & \textbf{SVHN} & \textbf{CIFAR10} & \textbf{SVHN}  \\
\hline
    
    0.95& 0.999&0.99&0.999& 0.99\\\hline
\end{tabular}
\end{table*}

\subsection{Effect of K-Neighbors on OOD Detection}
In this subsection, we explored the impact of varying the number of neighbors (k) on the AUROC.\\
To evaluate the sensitivity of our approach to neighboring parameters, we tested the entire out-of-distribution (OOD) pipeline using values of K = 5, 7, and 11 during the detection process. We obtained the same AUROC score of 0.999, with only a minor difference of $7 \times 10^{-6}$. This indicates that as long as the graph remains connected, varying the number of neighbors (K) within this range does not impact the performance of OOD detection. Therefore, we can fix K without requiring further hyperparameter tuning.
\begin{table}[ht!]
    \centering
    \caption{Overall classification accuracy as a function of percentile threshold $\tau$ applied to Mahalanobis distances.}
    \label{tab:threshold_vs_accuracy}
    \begin{tabular}{|c|c|}
        \hline
        \textbf{Percentile Threshold } & \textbf{Accuracy} \\ \hline 
   
        80  & 0.90 \\
        85  & 0.92 \\
        90  & 0.95 \\
        95  & 0.97 \\
        99  & 0.91 \\ \hline
        
    \end{tabular}
\end{table}
\subsection{Effect of Threshold on Accuracy}
While a threshold is often used for practical classification purposes, it is essential to distinguish between its role in decision-making and its role in overall performance evaluation. We tested different thresholds to evaluate their impact on accuracy as shown in Table \ref{tab:threshold_vs_accuracy}. The adjustment from 80 to 99 confirms that our detector's accuracy is significantly stronger in the exact specified area, with an optimal range identified between 85\% and 95\%. Accuracy improves from 80\% to 90\% and reaches 95.5\%. This increase reflects a better performance; however, when we push the threshold closer to 99\%, the detector becomes overly conservative; for instance, valid CIFAR-10 images begin to be rejected, causing the overall accuracy to drop to 91.5\%.

\end{document}